\documentclass[10pt, a4paper]{article}

\usepackage{lrec-coling2024} 

\usepackage{authblk}

\usepackage{xcolor}
\usepackage{hyperref}
\definecolor{darkblue}{rgb}{0, 0, 0.5}
\hypersetup{colorlinks=true, citecolor=darkblue, linkcolor=darkblue, urlcolor=darkblue}

\usepackage{nert} 
\usepackage{relsize}

\renewcommand{\nertcomment}[4]{\unskip}

\usepackage{listings}

\usepackage{natbib}
\usepackage{multibib}
\makeatletter
\def\@mb@citenamelist{cite,citep,citet,citealp,citealt,citepalias,citetalias}
\makeatother
\newcites{languageresource}{~}

\usepackage{graphicx}
\usepackage{tabularx}
\usepackage{soul}
\usepackage{linguex}

\usepackage{xstring}

\usepackage{color}

\makeatletter
\renewcommand\AB@affilsepx{\hspace{1em} \protect\Affilfont}
\makeatother

\definecolor{mblue}{HTML}{003F5C}
\definecolor{mviolet}{HTML}{58508D}
\definecolor{mpurple}{HTML}{BC5090}
\definecolor{msalmon}{HTML}{FF6361}
\definecolor{mgold}{HTML}{FFA600}

\title{Constructions Are So Difficult That Even Large Language Models Get Them Right for the Wrong Reasons}

\name{Shijia Zhou$^{1}$, Leonie Weissweiler$^{1,3}$, Taiqi He$^{2}$, \\ {\bf \large Hinrich Schütze$^{1,3}$, David R. Mortensen$^{2}$, Lori Levin$^{2}$}}

\address{ $^{1}$LMU Munich, 
$^{2}$LTI, Carnegie Mellon University,
$^{3}$Munich Center for Machine Learning\\ zhou.shijia@campus.lmu.de, weissweiler@cis.lmu.de}

\abstract{ 
In this paper, we make a contribution that can be understood from two perspectives: from an NLP perspective, we introduce a small challenge dataset for NLI with large lexical overlap, which minimises the possibility of models discerning entailment solely based on token distinctions, and show that GPT-4 and Llama 2 fail it with strong bias. We then create further challenging sub-tasks in an effort to explain this failure. From a Computational Linguistics perspective, we identify a group of constructions with three classes of adjectives which cannot be distinguished by surface features. This enables us to probe for LLM's understanding of these constructions in various ways, and we find that they fail in a variety of ways to distinguish between them, suggesting that they don't adequately represent their meaning or capture the lexical properties of phrasal heads. 
 \\ \newline \Keywords{LLMs, construction grammar, semantics, natural language inference, prompting} 
}

\begin{document}

\newcommand{\OCE}{OCE}
\newcommand{\LLC}{AAP}
\newcommand{\LLN}{EAP}
\newcommand{\CEC}{CEC}
\newcolumntype{L}{>{\raggedright\arraybackslash}X}
\newcolumntype{R}{>{\raggedleft\arraybackslash}X}

\maketitleabstract

\section{Introduction}\label{sec:intro}

\begin{figure}[t]
  \centering
  \begin{tikzpicture}[-stealth]
    \tikzset{word/.style={font=\relsize{-2.5}, text height=0.3cm, text depth=0.1cm},
      marked/.style={draw=red, rounded corners},
      label/.style={font=\relsize{-2.5}, text width=2cm},
      so adj/.style={draw=mblue},
      that/.style={draw=mpurple},
      verb/.style={draw=mpurple},
    }
    \node[matrix] (examples) {
      \node[label] (clausal comp) {Epistemic Adjective Phrase (EAP)}; &
      \node[word] (i3) {I}; &
      \node[word] (was3) {was}; &
      \node[word, marked, so adj, line width=1pt] (certain3) {\textbf{so} certain}; &
      \node[word, marked, verb, line width=1pt] (cried3) {\textbf{that} I saw you}; \\

      \node (blank) {}; & & &  \\
      \node[word] (blank) {\phantom{A}}; & & & \coordinate (abv so happy1); &  \coordinate (abv came1);\\
      \node[label] (clausal comp) {Affective Adjective Phrase (AAP)}; &
      \node[word] (i1) {I}; &
      \node[word] (was1) {was}; &
      \node[word, marked, so adj, line width=1pt] (so happy1) {\textbf{so} happy}; &
      \node[word, marked, verb, line width=1pt] (came1) {\textbf{that} I was freed}; \\
      
      \node (blank) {}; & & &  \\
      \node[word] (blank) {\phantom{A}}; & & & \coordinate (abv so happy2); &  \coordinate (abv cried2);\\
      \node[label] (clausal comp) {Causal Excess Construction (CEC)}; &
      \node[word] (i2) {It}; &
      \node[word] (was2) {was}; &
      \node[word, marked, so adj, line width=1pt] (so happy2) {\textbf{so} big}; &
      \node[word, marked, verb, line width=1pt] (cried2) {\textbf{that} it fell over}; \\
    };
    \draw (came1) -- (abv came1) -- node[label, above, midway] (cause1) {\ \ \ \ \ \ \ causality} (abv so happy1) -- (so happy1);
    \draw (so happy2) -- (abv so happy2) -- node[label, above, midway] (cause2) {\ \ \ \ \ \ \ causality} (abv cried2) -- (cried2);
  \end{tikzpicture}
  \caption[Example constructions]{Examples of the clausal complement and causal excess constructions. Constructions involving intensifier, adjective, and clausal complement}
  \label{fig:example-constructions}
\end{figure}
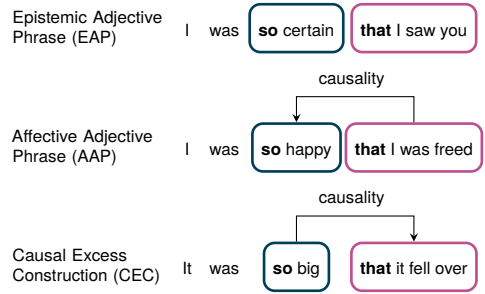

In this paper, we test the ability of large language models (LLMs) to identify different meanings in sentences that are superficially similar. The three sentences shown in Figure~\ref{fig:example-constructions} each contain the intensifier \textit{so}, an adjective that heads an adjective phrase, and a \textit{clausal complement}, a clause that is a dependent in the adjective phrase. In spite of their surface similarity, the three sentences have different semantic properties. In the first sentence, \textit{I was so certain that I saw you}, there are no causal connections between the adjective and the clausal complement. Seeing you did not make me certain, and being certain did not make me see you. In the second sentence, \textit{I was so happy that I was freed}, there is a causal connection. Being freed caused me to be happy. In the third sentence, \textit{It was so big that it fell over}, there is also a causal connection, but it is the reverse, being excessively big caused it to fall over. This is an example of the \textit{Causal Excess Construction} as studied by \citet{kay2012cleaning} and others. 

We examine these sentences from the perspective of Construction Grammar (CxG, ~\citealp{goldberg2006constructions, fillmore1988regularity, croft2001radical}), 
an approach to grammar in which arbitrary connections of form and meaning are not limited to the lexicon. In CxG meaning bearing units can contain multiple words, morphemes, and syntactic relations. 
The purpose of this paper is to test LLMs for their ability to differentiate between the three constructions in Figure~\ref{fig:example-constructions}, despite their surface similarity. 

The differences between the constructions can be observed semantically in terms of the presence of causality and the direction of causality as shown in Figure~\ref{fig:example-constructions}. The Causal Excess Construction (CEC) also differs from the others in underlying lexico-syntactic properties. This can be seen by removing the word \textit{so} from each sentence. \textit{I was certain that I saw you} and \textit{I was happy that I was freed} are grammatical sentences. In contrast, \textit{*It was big that it fell over} is not grammatical. The difference is explained in terms of \textit{licensing}. Every phrase or clause in a sentence must be licensed by a head that selects it. A simple example of licensing is that transitive verbs license direct objects (\textit{The student heard the lecture}) whereas intransitive verbs do not (\textit{*The student yawned the lecture}). In the examples we are dealing with here, epistemic and affective adjectives license complement clauses just as transitive verbs license direct objects. Adjectives like \textit{big} on the other hand, do not license clausal complements. In the CEC the clausal complement is licensed by the presence of \textit{so}, or more accurately, by the constellation of elements in the CEC construction \citep{kay2012cleaning}. Therefore the clausal complement cannot be present when \textit{so} is not present.

Further underlying syntactic properties of the three constructions make our work more interesting. Our experiments rely heavily on the fact that epistemic and affective adjectives can appear in the CEC as in \textit{I was so certain that I didn't plan for the alternative} and \textit{I was so happy that I cried}. In these sentences, the excess of certainty caused me not to plan for the alternative and the excess of happiness caused me to cry. Epistemic and Affective adjective constructions in the CEC may license two clausal complements, the first licensed by the adjective and the second licensed by the CEC: \textit{I was so certain that I was right that I didn't plan for the alternative}, \textit{I was so happy that I was freed that I cried}.

Importantly, for the experiments we present, any sentence without \textit{so} cannot be \CEC. So although \textit{so happy that I cried} can be \CEC, \textit{happy that I cried} cannot be \CEC ---crying is interpreted as the cause of happiness in the latter example.

In this work, we ask how well LLMs differentiate between affective adjective phrases (AAP), epistemic adjective phrases (EAP) and CEC. These offer an ideal testbed for linguistic probing, since they are structurally identical, with no exploitable surface cues. Above-baseline performance would have to stem from (1) knowledge about what each adjective can license, (2) knowledge about whether causality is typically associated with the adjective, and (3) understanding the direction of causality (adjective causes clause or clause causes adjective). Specifically, we investigate GPT-3.5 \cite{openai2021}, GPT-4 \cite{openai2023}, OpenAI's \texttt{ada-002} and Llama 2 \cite{touvron2023llama} with various questions, for each using both a prompt and a probing classifier. \footnote{See Appendix Table~\ref{tab:hyperparameters} for hyperparameters} We observe that LLMs exhibit limited capability to effectively discriminate between these constructions 
and display a strong bias towards CEC, meaning LLMs tend to judge sentences containing \textit{so... that...} as causal and the adjective being the reason for the clausal complement. Generally, Llama 2 demonstrates superior performance compared to both GPT-3.5 and GPT-4.

\section{Related Work}

Our work is situated in the framework of CxG \cite{croft2001radical, fillmore1988regularity, goldberg1995constructions, goldberg2006constructions, hoffmann2013oxford}. In this work, we define a construction as a pairing of form and meaning. We are considering AAP, EAP and CEC to be three different constructions. They differ in meaning (affective, epistemic, and excessive states) and in form (a clause that is licensed by a lexical head and a clause that is licensed by the intensifier \textit{so}).

Recent studies use CxG to probe the inner workings of LLMs \citep{weissweiler-etal-2022-better,chronis2023method, mahowald-2023-discerning}, and make general observations about the compatibility of the theory of CxG with the recent successes of LLMs \citep{goldberg2023chat, weissweiler-etal-2023-construction}. 
Most related to our work, \citet{mccoy-etal-2019-right} construct a challenging dataset for Natural Language Inference (NLI) that uses pairs of sentences with high lexical overlap. They show that the surface similarity of words almost always fools BERT into assuming that one sentence entails the other. Recent work \cite{si2022prompting, basmov2023chatgpt} suggests that the performance of recent LLMs on the \citet{mccoy-etal-2019-right} data has improved, though it is still far from perfect. We are therefore motivated to provide another challenging dataset to this line of inquiry.

\section{Dataset}
Our data collection process takes advantage of Universal Dependency \citep{nivre-etal-2020-universal} annotations, which we use for prefiltering before manual annotation. We use SPIKE \citep{shlain-etal-2020-syntactic} to access a parsed Wikipedia corpus and a parsed Amazon Reviews corpus. We establish that the parse trees of all CEC, \LLC\ and \LLN\ constructions have an edge labelled \texttt{ccomp} from the adjective to the head verb of the complement clause. 

We extract all sentences matching \textit{so... that...} pattern with a structural search string \texttt{The bookshelf was so adj:[tag]tall that it verb:[tag]toppled} in SPIKE, and group the sentences by adjective. Manually, where possible, we extract a sentence pair where one is CEC and one either \LLC\ or \LLN\, resulting in 111 such pairs. For the adjectives that cannot license a clausal complement, we extract 101 sentences, each with a different adjective. We call this class OCE (only Causal Excess). Our set of CEC sentences excludes any OCE adjectives.  

In total, we collect 323 sentences with 212 different adjectives.
\footnote{Data and code are available at \url{https://github.com/shijiazh/Constructions-Are-So-Difficult}}
An example of each sentence type is given in the first row of Table \ref{tab:transform}. 

\begin{table*}
\centering 
\tiny 
\begin{tabularx}{\textwidth}{llXXXXX}
\toprule
\textbf{Type} & \textbf{Transformation} 
& \textbf{{\OCE}} & \textbf{CEC} &  \textbf{{\LLC}} & \textbf{{\LLN}} \\
\midrule
\text{O} & Original & It was so big that it fell over. & I was so happy that I cried. & I was so happy that I was freed. & I was so certain that I saw you.\\
\midrule
\text{DS} & $-$ `so' & It was \{\} big that it fell over . & I was \{\} happy that I cried. & I was \{\} happy that I was freed. & I was \{\} certain that I saw you.\\
\text{DT} & $-$ `that' & It was so big \{\} it fell over . &  I was so happy \{\} I cried.  & I was so happy \{\} I was freed. & I was so certain \{\} I saw you. \\
\text{DST} & $-$ `so' \& `that' & It was \{\} big \{\} it fell over .  &  I was \{\} happy \{\} I cried. & I was \{\} happy \{\} I was freed. & I was \{\} certain \{\} I saw you. \\
\midrule
\text{AN} & $+$ `not'  & It was \textbf{not} so big that it fell over . &  I was \textbf{not} so happy that I cried. & I was \textbf{not} so happy that I was freed. & I was \textbf{not} so certain that I saw you.\\
\midrule
\text{P1} & main clause  & It was so big. & It was so happy. &  I was so happy. & I was so certain. \\
\text{P2} & sub. clause &  It fell over. & I cried. & I was freed. & I saw you.\\
\midrule
\text{Y-N} & yes–no question  & Did it fall over? &  Did I cry? & Was I freed? & Did I see you?\\
\bottomrule
\end{tabularx}
\caption{Transformations of the dataset with examples}
\label{tab:transform}
\end{table*}

\section{Natural Language Inference}
\label{subsection:entailment}

As shown in Table~\ref{tab:template1} (Prompts 1-1 to 1-4), we design four prompt variants to test whether LLMs can detect significant changes in entailment when \textit{so} is removed from \CEC\ sentences (DS type in Table \ref{tab:transform}). For example, \textit{I was so happy that I cried} does not automatically entail \textit{I was happy that I cried}, whereas \textit{I was so happy that I was freed} entails \textit{I was happy that I was freed}.

\begin{figure}[!ht]
\begin{center}
\centering
\includegraphics[width=\columnwidth]{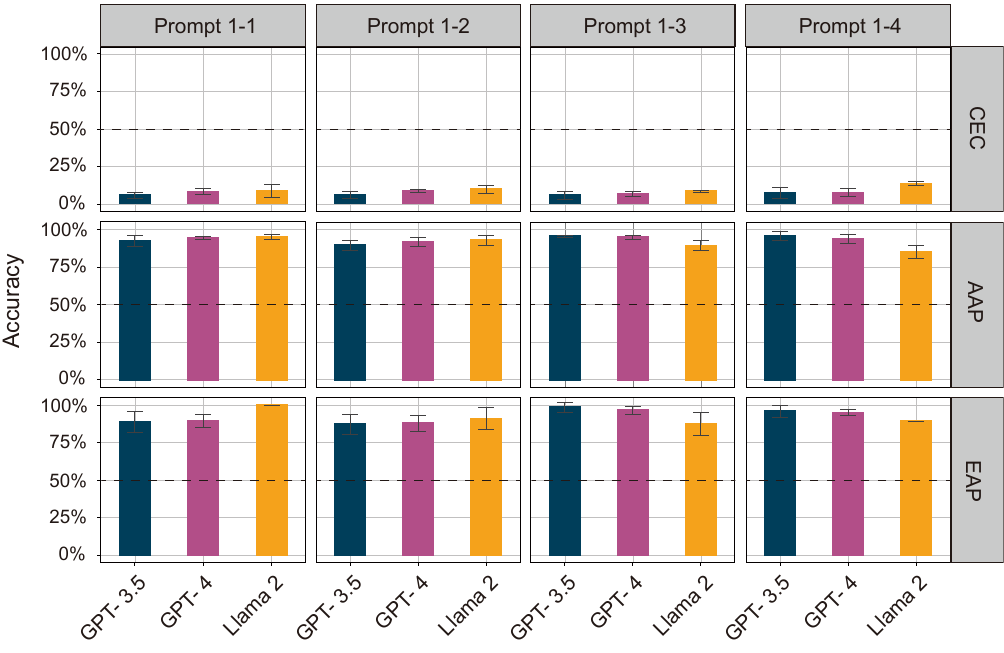} 
\caption{Performance of \CEC, \LLC, \LLN\  on the central NLI task. Corresponding gold labels are no entailment, no entailment, entailment, and entailment. All models have a strong bias to answer entailment. 
For OCE verbs, DS (delete so) is ungrammatical, therefore, we did not prompt on the OCE type.}
\label{fig:central}
\end{center}
\end{figure}

The results shown in Figure~\ref{fig:central} are striking: For CEC sentences, models achieve accuracy below 10\%, while demonstrating 90\% accuracy for AAP and EAP. It indicates a pronounced bias towards \textit{entailment} in the models, which replicates a behaviour shown for BERT: ``assume that a premise entails all hypotheses constructed from words in the premise'' by \citet{mccoy-etal-2019-right}. The following subsections investigate two hypotheses about why LLMs overestimate entailments: (i) LLMs are unable to identify causality in sentences (§\ref{subsec: Identifying Causality}); (ii) LLMs do not recognize the change in the direction of causality (§\ref{subsec: Direction of Causality}). \lw{insert section refs}

\begin{table*}
\centering 
\scriptsize 
\begin{tabularx}{\textwidth}{ccX} 
\toprule
\multicolumn{2}{c}{\textbf{{No.}}}& \textbf{{Template}} \\
\midrule
\multirow{4}{*}{\centering 1}&1 &  premise: \textbf{O} \textbackslash n hypothesis: \textbf{DS}  \textbackslash n Classify as entailment, no entailment, or contradiction. \\
&2 &  premise: \textbf{O}  \textbackslash n hypothesis: \textbf{DST}  \textbackslash n Classify as entailment, no entailment, or contradiction. \\
&3 &  \textbf{O}  Can we infer that ``\textbf{DS}"? \textbackslash n Answer with yes, no or uncertain.  \\
&4 &  {\textbf{O}  Can we infer that ``\textbf{DST}"? \textbackslash n Answer with yes, no or uncertain.}  \\ 
\midrule
\multirow{2}{*}{\centering 3}&1 & \textbf{O}  \textbackslash n Is there a causal relationship between the main clause and the subordinate clause? \textbackslash n Answer with yes, no or uncertain. \\
&2 & \textbf{O}  \textbackslash n Part1: \textbf{P1}  \textbackslash n Part2: \textbf{P2}  \textbackslash n Is there a causal relationship between part 1 and part 2? \textbackslash n Answer with yes, no or uncertain. \\
\midrule
\multirow{5}{*}{\centering 4}&1 & premise: \textbf{O}  \textbackslash n hypothesis: \textbf{P2}  \textbackslash n Classify as entailment, no entailment, or contradiction. \\
&2 & \textbf{AN}  \textbackslash n \textbf{Y-N} \textbackslash n Answer with yes, no or uncertain. \\
&3 &  \textbf{O}  \textbackslash n Part1: \textbf{P1}  \textbackslash n Part2: \textbf{P2}  \textbackslash n Can we infer that Part1 is the cause of Part2? \textbackslash n Answer with yes, no or uncertain. \\
&4 &  \textbf{O}  \textbackslash n Part1: \textbf{P1}  \textbackslash n Part2: \textbf{P2}  \textbackslash n Can we infer that Part2 is the cause of Part1? \textbackslash n Answer with yes, no or uncertain. \\
&5 &  \textbf{O}  \textbackslash n This entails one of two options. \textbackslash n 1) \textbf{P1}  because \textbf{P2}  \textbackslash n 2) \textbf{P2}  because \textbf{P1}  \textbackslash n Answer with the correct number.\\
\bottomrule
\end{tabularx}
\caption{Prompt templates of all tasks. To test the stability of model responses, we design variants of each prompt, removing only \textit{so} from premise as hypothesis or removing both \textit{so} and \textit{that}.} \drm{Fix this table}
\label{tab:template1}
\end{table*}

\subsection{Experimental Setup}
We conduct our experiments with two methods. The first approach involves the development of both implicit and explicit prompts. In the second approach, we extract the last-layer embeddings from LLMs and then apply perceptron-based classification to these embeddings, to assess how well the CEC, OCE, AAP and EAP categories
are internally represented in the models.

We employ a perceptron classifier to test the final layer embeddings of Llama 2, and \texttt{ada-002} sentence embeddings. For Llama 2, we use the mean over token embeddings as a sentence embedding. We hypothesise that the contextual representation of the adjective itself will encode the presence or direction of causality, and therefore also test its embedding separately.
When the two input classes are imbalanced, we upsample the smaller class. We group sentences containing the same adjectives together and train the perceptron using cross-validation over adjectives, meaning that the adjective itself is no longer an exploitable feature. A Bag-of-Words (BoW) model serves as the baseline.

\subsection{Identifying Causality}
\label{subsec: Identifying Causality}

\paragraph{Prompting} 

As depicted in Table~\ref{tab:template3}, we devise two prompts to assess the models' capability to identify causal relationships. The first simply asks about a causal relationship between the main and the subordinate clause, while the second additionally provides the pre-segmented parts.
In the \LLN\ category, all models have accuracy below 20\%, suggesting a predisposition to infer causality in sentences containing \textit{so... that...} even when the adjective is epistemic. Llama 2 displays a stronger bias, attributing causality to over 90\% of sentences in all categories. \drm{Llama 2 is superstitious (if we are attributing cognitions to them now).}
Combining the previous observation that Llama 2 tends to categorise every sentence containing \textit{so} and \textit{that} as grammatically correct, along with its sensitivity to the absence of \textit{so} in CEC, this suggests a limited grasp of semantic nuances and an overreliance on simple lexical cues.
GPT models both struggle about equally, with less than 50\% accuracy in \LLC\ instances. Even more perplexing, EAP sentences are classified as causal at a significantly higher rate than CEC and OCE.

\begin{table}
\centering 
\scriptsize 
\begin{tabularx}{\columnwidth}{clXXXXc} 
\toprule
\textbf{No.} & \textbf{Model} & \textbf{\OCE} & \textbf{\CEC} &  \textbf{\LLC} & \textbf{\LLN} & \textbf{Gold Lab.}  \\
\midrule
\multirow{3}{*}{3-1} & GPT-3.5 & 67.33 & 60.90 & 41.68 & \textbf{18.57}  & \multirow{3}{*}{Y|Y|Y|N} \\
& GPT-4 & 63.37 & 58.74 & 41.20 & 15.00  &  \\
& Llama 2  & \textbf{95.05} & \textbf{98.65} & \textbf{95.18} & 08.93 &\\ \midrule
\multirow{3}{*}{3-2} & GPT-3.5 & 54.46 & 64.14 & 49.15 & 06.43  & \multirow{3}{*}{Y|Y|Y|N}\\
& GPT-4 & 57.03 & 65.95 & 46.02 & 04.28 &  \\
& Llama 2 & \textbf{95.54} & \textbf{99.10} & \textbf{92.78} & \textbf{08.93}  &  \\
\bottomrule
\end{tabularx}
\caption{Accuracy of the task of identifying causality with 
different prompts}
\label{tab:template3}
\end{table}

\begin{figure}[!ht]
\begin{center}
\centering
\includegraphics[width=\columnwidth]{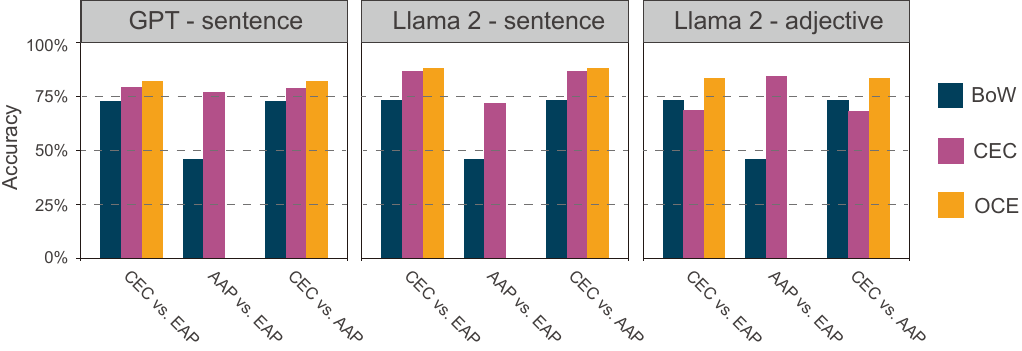} 
\caption{ Accuracy of perceptrons trained with different embeddings across three tasks. In all sub-tasks involving CEC structures, we attempt to replace CEC with OCE. OCE adjectives are mutually exclusive with those in EAP and AAP.}
\label{fig:perceptron}
\end{center}
\end{figure}

\paragraph{Probing Classifier} \label{sec: idpc}
The classifiers for \CEC\ vs \LLN\ and \LLC\ vs \LLN\ serve to assess the models' representation of causality. 
As illustrated in Figure~\ref{fig:perceptron}, on \CEC\ vs \LLN, perceptrons trained on sentence embeddings beat the baseline while those trained on adjective embeddings do not, suggesting that causality is encoded, but not necessarily in the adjective. \drm{This argument could be stated more clearly.}
Interestingly, the adjective perceptron beats the baseline on the \LLC\ vs \LLN\ test, even though the sets of adjectives are mutually exclusive and we perform cross-validation over them, meaning that the only source of information left would be a deeper commonality between them. This suggests that the models may have learned common features for affective and epistemic adjectives, respectively. Furthermore, the result also demonstrates that the distinction between EAP and CEC is more pronounced in the model's perspective compared to the distinction between EAP and AAP, especially for Llama 2 sentence embeddings.

\subsection{Direction of Causality}
\label{subsec: Direction of Causality}

\paragraph{Prompting}

The negator \textit{not} before \textit{so} influences the truth value of the subclause for \CEC\ but has no influence in \LLC. For instance, \textit{He was not so big that he fell} (CEC) does not imply that he fell, while \textit{He was not so happy that he went.} (\LLC) suggests that he went. Asking the model to distinguish between these is equivalent to distinguishing the direction of causality.

Therefore, we devise an explicit NLI prompt 4-1 \citep{webson-pavlick-2022-prompt} and an implicit prompt 4-2 (AN + Y-N) to explore these effects. Accurate answers depend on the models' precise understanding of the causal direction. We also introduce prompt 4-3, where P1 and P2 are provided, and the question is whether P1 is the cause of P2. Prompt 4-4 maintains the same structure but now inquires whether P2 is the cause of P1. Prompt 4-5 is structured as a multiple-choice question between the two directions.

As can be seen in Table~\ref{tab:direction}, results for prompts 4-1 and 4-2 suggest that they bias all models towards answering \textit{yes} for any sentence. By contrast, prompts 4-3 and 4-4 show a clearer picture, still biased, but Llama 2 scores clearly correlate with the gold label.
By comparing these two sets of prompts, we can discern that Llama 2's responses are grounded in an analysis of P1 and P2, rather than simply providing uniform \textit{yes} or \textit{no} answers. Interestingly, prompt 4-5 elicit better performance from the GPT models in contrast, suggesting that it might be more suited to the multiple-choice answer format. We conclude that all models have some representation of the direction of causality, but it is far from perfect.

\begin{table}
\centering 
\scriptsize 
\begin{tabularx}{\columnwidth}{cLrrrc} 
\toprule
\textbf{{Type}} & \textbf{{Model}} & \textbf{{\OCE}} & \textbf{{\CEC}} & \textbf{{\LLC}}  & \textbf{{Gold Label}}  \\
\midrule
\multirow{3}{*}{4-1} & GPT-3.5 & 28.71 & 29.19 & 62.41 & \multirow{3}{*}{N|N|Y} \\
& GPT-4 & 26.93 & 29.01 & \textbf{62.65} &  \\
& Llama 2 & \textbf{39.60} & \textbf{49.10} & 53.62 &\\
\midrule
\multirow{3}{*}{4-2} & GPT-3.5 & 4.55 & 2.52 & 60.24 & \multirow{3}{*}{N|N|Y}\\
& GPT-4 & 4.95 & 2.16 & \textbf{60.72} &  \\
& Llama 2 & \textbf{16.34} & \textbf{18.47} & 40.97 &  \\
\midrule
\multirow{3}{*}{4-3} & GPT-3.5 & 74.46 & 82.88 & 13.25 & \multirow{3}{*}{Y|Y|N}\\
& GPT-4 & 77.03 & 83.96 & 8.91 &  \\
& Llama 2 & \textbf{87.13} & \textbf{93.69} & \textbf{46.99} &  \\
\midrule
\multirow{3}{*}{4-4} & GPT-3.5 & 50.69 & 42.70 & 44.09 & \multirow{3}{*}{N|N|Y}\\
& GPT-4 & 48.31 & 40.18 & 47.95 &  \\
& Llama 2 & \textbf{77.23} & \textbf{71.17} & \textbf{81.92 }&  \\
\midrule
\multirow{3}{*}{4-5} & GPT-3.5 & \textbf{61.19} & \textbf{60.72} & 77.59 & \multirow{3}{*}{2)|2)|1)}\\
& GPT-4 & 60.80 & 54.78 & \textbf{79.27} &  \\
& Llama 2 & 51.98 & 45.49 & 78.31 &  \\
\bottomrule
\end{tabularx}
\caption{Accuracy of direction of causality task with different prompts. Y: yes/entailment, N: no/contradiction.}
\label{tab:direction}
\end{table}

\paragraph{Probing Classifier}

The classifier for \CEC\ vs \LLC\ serve to assess the models' capability to differentiate the direction of causality. The results in Figure~\ref{fig:perceptron} are similar to those for identifying causality in Section~\ref{sec: idpc}, with the notable exception of the \OCE\  test set, which is easiest for the adjective classifiers with no obvious explanation.

Figure~\ref{fig:perceptron} displays that on CEC vs AAP, similar to CEC vs EAP, models trained with sentence embeddings outperform the baseline, while those trained with adjective embeddings slightly lag behind. Additionally, the perceptron attains the highest accuracy on the OCE test set. 

As adjectives in \OCE\ can only appear in \CEC, while adjectives in CEC can also occur in \LLC\ or \LLN, this can be interpreted as \OCE's adjective embeddings being more easily identified as belonging to the \CEC\ structure than those of CEC.

\section{Grammatical Acceptability}

We perform experiments on grammatical acceptability to see whether LLMs are sensitive to the difference between embedded clauses that are lexically licensed by an adjective and embedded clauses that are licensed by the CEC construction.  
Following~\citet{mahowald-2023-discerning}, we prompt LLMs for grammaticality judgements by first presenting eight pairs of sentences from the CoLA corpus \citep{warstadt-etal-2019-neural}, which consists of minimal pairs of grammatical and ungrammatical sentences extracted from linguistics literature. Then the target sentence is inserted, and the model is asked to generate one token, \textit{good} or \textit{bad}. 

\begin{table}
\centering 
\scriptsize 
\begin{tabularx}{\columnwidth}{llXXXXc} 
\toprule
\textbf{{No.}} & \textbf{{Model}} & \textbf{{\OCE}} & \textbf{{\CEC}} & \textbf{{\LLC}} &  \textbf{{\LLN}}  &  \textbf{{Gold Label}}  \\
\midrule
\multirow{3}{*}{O} & GPT-3.5  & 89.31 & 92.43 & 80.96 & 73.57 & \multirow{3}{*}{G|G|G|G} \\
& GPT-4 & 89.31 & 92.97 & 81.93 & 73.57 &  \\
& Llama 2 & \textbf{100.00} & \textbf{100.00} & \textbf{100.00} & \textbf{100.00} &  \\
\midrule
\multirow{3}{*}{DS} & GPT-3.5 & 36.83 & \textbf{84.69} & 87.23 & 79.28 & \multirow{3}{*}{B|G|G|G} \\
& GPT-4 & 36.43 & 84.32 & \textbf{88.92} & 78.57 &  \\
& Llama 2 & \textbf{39.58} & 76.30 & 80.62 & \textbf{83.34} &  \\
\midrule
\multirow{3}{*}{DT} & GPT-3.5 & 80.40 & 89.55 & 72.29 & 60.00 & \multirow{3}{*}{G|G|G|G} \\
& GPT-4 & 80.40 & 88.83 & 70.60 & 56.43 &  \\
& Llama 2 & \textbf{100.00} & \textbf{100.00} & \textbf{100.00} & \textbf{100.00} &  \\
\midrule
\multirow{3}{*}{DST} & GPT-3.5  & \textbf{57.83} & \textbf{67.39} & 67.95 & 65.71 & \multirow{3}{*}{B|G|G|G} \\
& GPT-4 & \textbf{57.83} & 66.67 & \textbf{68.92} & 65.72 &  \\
& Llama 2 & 49.77 & 49.97 & 60.50 & \textbf{76.46} &  \\
\midrule
\multirow{3}{*}{AN} & GPT-3.5 & 83.76 & 90.63 & 74.46 & 69.29 & \multirow{3}{*}{G|G|G|G} \\
& GPT-4 & 84.36 & 90.45 & 76.14 & 74.29 &  \\
& Llama 2 & \textbf{100.00} & \textbf{100.00} & \textbf{100.00} & \textbf{100.00} &  \\
\bottomrule
\end{tabularx}
\caption{Accuracy of the grammaticality task. Bold font indicates the models with the highest accuracy for a type and transformation. G: good, B: bad.}
\label{tab:grammaticality}
\end{table}

We test the original sentences (O) and four transformations DS, DT, DST, and AN, as mentioned in Table~\ref{tab:transform}. 
Deleting \textit{that} (DT) or adding \textit{not} (AN) will not affect the grammaticality of CEC, EAP and AAP, whereas removing \textit{so} (DS) from OCE makes them ungrammatical.

As can be seen in Table~\ref{tab:grammaticality}, compared to \CEC\, \OCE\  is more likely to be rated as \textit{bad} by both GPT-3.5 and GPT-4, regardless of the transformation. 

It demonstrates that GPT models indeed detect the distinction between \OCE\  and \CEC\, especially regarding their reliance on the grammaticality with \textit{so}. 
Notably, for GPT models, removing \textit{that} from \LLC\ and \LLN\ sentences results in more \textit{good}, whereas removing \textit{that} from DS sentences tends to yield more \textit{bad} ratings even though it has no effect on grammaticality.

Llama 2's answers deviate from that of GPT models. It rates all O, DT, and AN sentences as \textit{good}, which is exactly the gold label, signifying its robust inclination to not only consider sentences featuring \textit{so... that...} as acceptable,\drm{acceptable? well-formed?} but also acknowledge the possibility of omitting \textit{that} in such contexts. 
However, its performance on DS and DST is far from perfect. 

Additionally, GPT4 performs better on the DS variants of AAP.\footnote{A probing classifier for this prompting task would not be well-defined.}

\section{Conclusion}

Overall, our most striking result remains that no LLM performed adequately on our NLI task, and that this result is not sufficiently explained by the mediocre but better-than-baseline performance on the sub-tasks. Llama 2 performed better in those than the GPT models, but generally, prompting results are often consistently below random and probing classifier results only slightly above baseline.
Interestingly, GPT-4 does not perform significantly better than GPT-3.5 at any task. 

Both in the central NLI task, and the sub-tasks, all LLMs show bias to offer positive answers.
Llama 2 demonstrates a more comprehensive understanding of the grammatical structures within CEC, an enhanced ability to identify causality within sentences, and a greater proficiency in ascertaining the direction of causality compared to GPT models. These findings align with our initial observations in the central NLI task. \drm{These are interesting observations. Is it possible to provide some statistical tests to determine whether these differences are larger that you would expect from chance?}

We exclude the following from the current work: (1) Extraposition from clausal subject (\textit{That I left was so bad}/\textit{It was so bad that I left}) (2) CEC meanings with other intensifiers such as \textit{enough} and non-finite clauses (\textit{big enough to fall over}) (3) CEC headed by nouns (\textit{so many people that the police came}). We have also not investigated the CEC in sentences other than copula sentences, or when the CEC adjective is part of a noun phrase (\textit{I met many people so short that they couldn't reach the counter}).

\newpage

\section*{Bibliographical References}\label{sec:reference}

\bibliographystyle{lrec_natbib}
\bibliography{custom,anthology}

\appendix

\section*{Appendix}

\begin{table}
\begin{tabular}{l}
\toprule
Now we are going to say which sentences are \\
acceptable (i.e., grammatical) and which are not.                  \\
                                                     \\
Sentence: Flosa has often seen Marn.                               \\
Answer: good                                                       \\
\\
Sentence: Chardon sees often Kuru.                                 \\
Answer: bad                                                        \\
\\
Sentence: Bob walk.                                                \\
Answer: bad                                                        \\
\\
Sentence: Malevolent floral candy is delicious.                    \\
Answer: good                                                       \\
\\
Sentence: The bone chewed the dog.                                 \\
Answer: good                                                       \\
\\
Sentence: The bone dog the chewed.                                 \\
Answer: bad                                                        \\
\\
Sentence: I wonder you ate how much.                               \\
Answer: bad                                                        \\
\\
Sentence: The fragrant orangutan sings loudest at                  \\
Easter.                                                            \\
Answer: good                                                       \\
\\
Sentence: {[}TEST SENTENCE GOES HERE{]}                            \\
Answer: \\                                           \bottomrule          
\end{tabular}
\caption{Few-shot CoLA prompts template created by \citet{mahowald-2023-discerning}. We tested 5 types of sentence: O, DS, DT, DST and AN.}
\label{tab:CoLA}
\end{table}

\begin{table}
\begin{tabular}{lrr}
\toprule
                   & \textbf{GPT-3.5/GPT-4} & \textbf{Llama 2} \\ \midrule
temperature        & 1             & 0.7     \\
top\_p             & 1             & 0.95    \\
top\_k             & -             & 40      \\
max\_tokens        & null          & 512     \\
frequency\_penalty & 0             & -       \\
presence\_penalty  & 0             & -      \\ \bottomrule
\end{tabular}
\caption{Hyperparameters for GPT-3.5, GPT-4, OpenAI's ada-002 and Llama 2. For each prompt, we repeat the mean over six runs of the experiment.}
\label{tab:hyperparameters}
\end{table}


\begin{table*}[p]
\small{
\begin{tabularx}{\textwidth}{llL}
\toprule
{ \textbf{Adjective}} &
  \textbf{Type} &
  \textbf{Sentence} \\ \midrule
{ }                             & CEC & In one XFM show , he became so frustrated that he left the room before Karl finished the segment .    \\
\multirow{-2}{*}{{ frustrated}} & AAP & I am so frustrated that a \$ 500 purchase brought such short lived joy .                              \\ \midrule
{ }                             & CEC & Mandhata had dominated the whole planet and he became so proud that he wanted to rule heaven   also . \\
\multirow{-2}{*}{{ proud}} &
  AAP &
  My dad was so proud that his son made " aliyah " . \\ \midrule
 &
  CEC &
  One man was so afraid that he camped in the middle of his flock , hoping to evade   patrolling cowboys . \\
\multirow{-2}{*}{afraid} &
  EAP &
  He was so afraid   that rival loyalist inmates wished to kill him inside the prison . \\ \midrule
 &
  CEC &
  Like Napoleon , Hitler was so optimistic that he falsely believed he 'd make it to Moscow before Winter   . \\ 
\multirow{-2}{*}{optimistic} &
  EAP &
  I am so optimistic that I made the best choice . \\ \midrule
abrupt &
  OCE &
  The growth was so abrupt that a village sprang . \\ \midrule
beautiful &
  OCE &
  The palace was so beautiful that the king of Mengwi heard of Tan Cin Jin . \\ \bottomrule
\end{tabularx}
}
\caption{Examples from the collected database. CEC represents causal excess construction, where the adjective is interpreted as the cause of the complement. AAP stands for affective adjective phrases, which usually trigger an inference that the complement caused the feeling expressed by the adjective. EAP stands for epistemic adjective phrases,  which lexically liscence non-causal complement.}
\label{tab:sentence pairs}
\end{table*}

\end{document}